\documentclass[letterpaper, 10 pt, conference]{ieeeconf}  

\IEEEoverridecommandlockouts                              

\usepackage{graphics} 
\usepackage{epsfig} 
\usepackage{amsmath} 
\usepackage{amssymb}  
\usepackage{import}
\usepackage{hhline}
\usepackage{color}
\usepackage[table]{xcolor}

\usepackage{lipsum} 
\usepackage{multirow}
\usepackage[font=small]{caption}

\title{\LARGE \bf
Towards Feasible Dynamic Grasping: Leveraging Gaussian Process Distance Field, SE(3) Equivariance, and Riemannian Mixture Models}

\author{Ho Jin Choi$^{\dagger}$ and Nadia Figueroa$^{\dagger}$
\thanks{$^{\dagger}$Ho Jin Choi and Nadia Figueroa are with the Department of Mechanical Engineering and Applied Mechanics, School of Engineering and Applied Science, University of Pennsylvania, Pennsylvania, PA 19104 USA (email: \{cr139139,nadiafig\}@seas.upenn.edu)}
}
\begin{document}
\maketitle
\thispagestyle{empty}
\pagestyle{empty}

\begin{abstract}
This paper introduces a novel approach to improve robotic grasping in dynamic environments by integrating Gaussian Process Distance Fields (GPDF), SE(3) equivariant networks, and Riemannian Mixture Models. The aim is to enable robots to grasp moving objects effectively. Our approach comprises three main components: object shape reconstruction, grasp sampling, and implicit grasp pose selection. GPDF accurately models the shape of objects, which is essential for precise grasp planning. SE(3) equivariance ensures that the sampled grasp poses are equivariant to the object's pose changes, enhancing robustness in dynamic scenarios. Riemannian Gaussian Mixture Models are employed to assess reachability, providing a feasible and adaptable grasping strategies. Feasible grasp poses are targeted by novel task or joint space reactive controllers formulated using Gaussian Mixture Models and Gaussian Processes. This method resolves the challenge of discrete grasp pose selection, enabling smoother grasping execution. Experimental validation confirms the effectiveness of our approach in generating feasible grasp poses and achieving successful grasps in dynamic environments. By integrating these advanced techniques, we present a promising solution for enhancing robotic grasping capabilities in real-world scenarios.

\end{abstract}
\section{Introduction}
\label{sec:intro}
Grasping is a fundamental action enabling robots to execute manipulation tasks. The difficulty in grasping relies on finding an optimal grasp pose for the robot to perform, i.e., grasp sampling. To accomplish effective grasp sampling, consideration of three major components is required: (a) the shape and physical properties of the object; (b) the robot's kinematics; and (c) potential collisions with the surroundings. Therefore, research in this field focuses on gathering information about these factors to help robots find good grasp positions.

In real-world scenarios, accurately determining object shapes can be hard due to occlusions or limited information. Analytic grasp samplers use force closure analysis \cite{nguyen1988constructing} or differentiable simulation \cite{Turpin2023FastGraspDDM} to find good grasp poses, but they rely on knowing the object's shape, which isn't always possible. To address this limitation, new approaches have emerged in diverse manners, all without needing detailed prior knowledge about object shapes. These approaches include fitting shape primitives to an object's partial point cloud \cite{Wu2023LearningFreeGO, Kim2023DSQNetAD}, estimating the composition of multiple shape primitives using neural networks (NN) \cite{Lin2019UsingSD}, completing object shape using NN \cite{Varley2016ShapeCE}, or employing encoder-decoder architectures to implicitly learn object shape \cite{Mousavian20196DOFGV}. These methods hold potential but often address only certain aspects of object shape, lacking vital information for effective control. Surface details and distance functions are critical for stable grasping and avoiding collisions. Recent efforts aim to learn the complete signed distance function (SDF) \cite{Urain2022SE3DiffusionFieldsLS}, which aids in filtering and validating data. Additionally, distance functions assist in smooth gripper operation and enable strategies for re-grasping if the object shifts during manipulation.

\begin{figure}[!tbp]
  \centering
  \includegraphics[width=\linewidth]{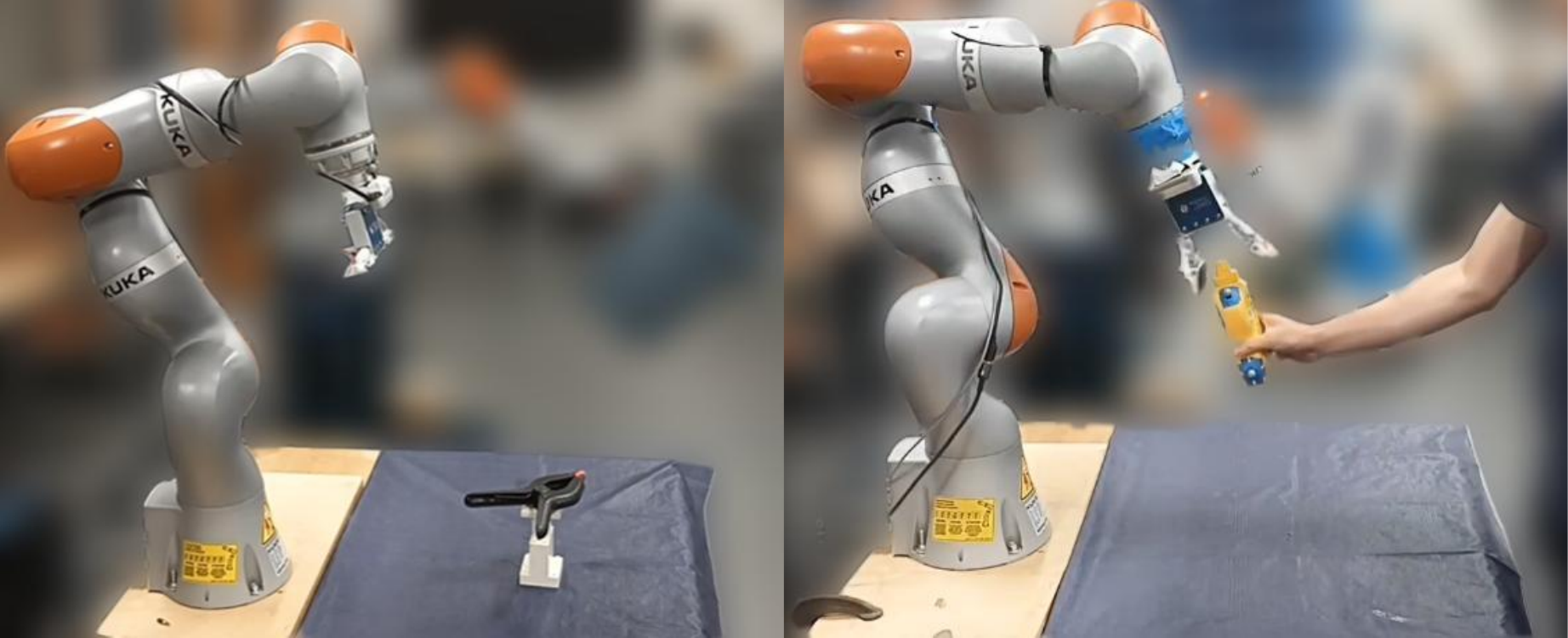}
  \vspace{-15pt}
  \caption{Experimental setup used to evaluate our feasible dynamic grasping framework (left: static object, right: dynamic object). An Intel RealSense camera mounted on the robot's end-effector is used to obtain a partial point cloud of the object and its pose (optitrack is used to track the object in dynamic scenarios) with no prior knowledge about the object category. \label{realrobot}}
  \vspace{-20pt}
\end{figure}
When addressing the robot's kinematic constraints and collisions, two distinct approaches emerge: the explicit sampling of grasp poses achieved by learning kinematics and collisions implicitly \cite{Sundermeyer2021ContactGraspNetE6}, and the transformation of the grasp sampling challenge into a multi-objective optimization through the acquisition of a grasp pose cost or distance function \cite{Weng2022NeuralGD, Urain2022SE3DiffusionFieldsLS}. The first method is computationally fast but limited to specific situations, hindering its applicability across diverse scenarios. In contrast, the latter approach is more adaptable, as it can consider different constraints in the control optimization. It does not suffer from the challenges of grasp selection problem and can be applied to dynamic objects, unlike the linear bandit method \cite{Wang2019ManipulationTO} for static objects. However, extending this approach to task-specific grasp sampling, accounting for affordances, presents difficulties. Furthermore, SE(3) grasp pose distances don't necessarily correlate with the nearest pose in joint space.
\begin{figure*}[!tbp]
  \includegraphics[width=\textwidth]{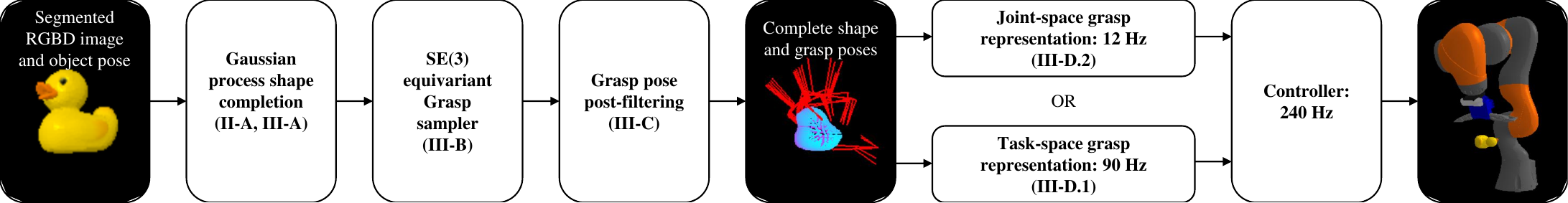}
  \caption{\textbf{Feasible Dynamic Grasping Framework} pipeline. Given a partial point cloud of an object and its pose, we reconstruct the shape using the Gaussian Process Distance Field (Section \ref{sec:gpis}) with a refined distance metric as in Section \ref{sec:distance_field}. Next, we sample poses from an SE(3) equivariant grasp sampler (Section \ref{sec:grasp_sampler}). Finally, the discrete set of feasible grasp poses is used to i) construct a continuous Riemannian GMM for dynamic task-space control (Section \ref{sec:task_control}) or ii) convert to joint-space via IK to construct a continuous GP for dynamic control in joint-space (Section \ref{sec:joint_control}). The grasp sampling and controller components can operate asynchronously, allowing for real-time adaptation. \label{pipeline}}
\vspace{-15pt}
\end{figure*}

In scenarios involving human-robot interaction (HRI), it is crucial to address dynamic objects and implement reactive control strategies \cite{dsbook}. One viable approach is to optimize the end-effector \cite{Kim2014CatchingOI} or sampled trajectory \cite{Akinola2021DynamicGW} based on the object's pose. Recent research has explored leveraging the consistency of learned grasp latent vectors to match grasp poses across time steps \cite{Liu2023TargetreferencedRG, Fang2022AnyGraspRA}. Furthermore, some use RL for object handover control policies \cite{Christen2023LearningHH}. While these methods show promise in learning the dynamics of grasping and reducing modeling complexity, their direct applicability to diverse scenes or methods remains a challenge.

\textbf{Contribution} This work introduces a sequential modular approach for sampling 6D grasp poses, reactive motion planning, and controlling dynamic objects (Figure \ref{pipeline}). Our pipeline assumes a segmented point cloud of a rigid object, simplifying initial modeling. It consists of three main steps:
\begin{enumerate}
    \item \textbf{Object Shape Reconstruction}: Reconstruct the object's shape using Gaussian Process distance field \cite{Gentil2023AccurateGP}, providing an approximate SDF for resampling the complete point cloud for comprehensive representation.
    \item \textbf{SE(3) Equivariant Grasp Sampling and Filtering}: Perform computationally efficient SE(3) equivariant grasp sampling and post-filtering with Gaussian Mixture Model-based reachability analysis for enhancing grasp pose quality and robustness.
    \item \textbf{Implicit Grasp Pose Selection and Control}: Control the robot towards the implicit grasp pose using gradients from probability or distance functions, employing either task space or joint space methods for smooth motion to do automatic grasp selection.
\end{enumerate}
By integrating these steps, our pipeline offers an effective solution for generating and evaluating grasp poses and facilitating motion planning and control in dynamic environments (Figure \ref{realrobot}).


\section{Preliminaries}
Next we described the probabilistic and mathematical tools we use to build our feasible dynamic grasping framework.
\subsection{Gaussian Process Distance Field (GPDF)}
\label{sec:gpis}
Gentil et al. \cite{Gentil2023AccurateGP} proposed using Gaussian processes (GPs) to estimate an object or entire scene's distance field solely from its surface points. Let $\mathbf{x}\in \mathbb{R}^m$ be a point and $\mathbf{x}'$ be points lying on the surface $\mathcal{S}\in\mathbb{R}^m$. The occupancy field $o(\mathbf{x})\in\mathbb{R}$ is modeled as $o(\mathbf{x})\sim\mathcal{GP}(0,k_o(\mathbf{x},\mathbf{x}'))$, where $k_o(d)$ is the covariance kernel. The inferred occupancy from point $\mathbf{x}$ is calculated as:
\begin{equation}
\hat{o}(\mathbf{x})=k_o(\mathbf{x},\mathbf{X})(K_o(\mathbf{X},\mathbf{X})+\sigma_o^2 I)^{-1}\mathbf{1}
\end{equation}
where $K_o(\mathbf{X},\mathbf{X})$ is the covariance matrix of the observed surface points $\mathbf{X}$, $\sigma_o^2$ is the noise variance, and $\mathbf{1}$ is a vector of ones. Define the inverse function $r$ as $r(k_o(\mathbf{x}, \mathbf{x}'))=||\mathbf{x}-\mathbf{x}'||=d$. Then, the distance field between $\mathbf{x}$ and the surface $\mathcal{S}$ is given by:
\begin{equation}
\hat{d}(\mathbf{x})=r(\hat{o}(\mathbf{x}))
\end{equation}
This can be understood as reversing the effect of the kernel function, which essentially represents the distance between two points when there is only one surface point and a query point involved. The gradient of the distance field, useful for obtaining surface normals, is analytically derived as:
\begin{equation}
    \nabla d(\mathbf{x})=\frac{\partial r}{\partial o}\left(\nabla k_o(\mathbf{x},\mathbf{X})(K_o(\mathbf{X},\mathbf{X})+\sigma_o^2 I)^{-1}\mathbf{1}\right)
\end{equation}
GPs provide an uncertainty measure expressed as:
\begin{equation}
    \begin{aligned}
        \mathbf{var}(o(\mathbf{x})) = k_o(\mathbf{x},\mathbf{x})\\
        - k_o(\mathbf{x},\mathbf{X})(K_o(\mathbf{X},\mathbf{X})+\sigma_o^2 I)^{-1}k_o(\mathbf{X},\mathbf{x})
    \end{aligned}
\end{equation}
This measure can assess the certainty of surface points, aiding in shape completion, as detailed in Section \ref{sec:distance_field}. This SDF representation and uncertainty give us other measures to further filter out good grasp poses.

\subsection{SE(3) equivariance}
\label{sec:se3}
The SE(3) Lie group is commonly used in robotics, especially for describing end-effector and grasp poses. It provides a unified representation consisting of rotation $R\in SO(3)$ and translation $t\in\mathbb{R}^3$. A point in this representation is denoted as $T=\big[\begin{smallmatrix} R & t\\ 0 & 1 \end{smallmatrix}\big] \in SE(3)$. For grasp samplers focusing on object-centric approaches \cite{Simeonov2021NeuralDF, Sen2023SCARP3S}, SE(3) equivariance is crucial. This ensures that feasible grasp poses of an object remain consistent relative to the object's pose. Let $f:\mathbb{R}^{3\times n}\rightarrow SE(3)$ be a grasp sampler, where $\mathbf{P}\in\mathbb{R}^{3\times n}$ represents the object's point cloud and $G=\big[\begin{smallmatrix} R_G & t_G\\ 0 & 1 \end{smallmatrix}\big] \in SE(3)$ is a grasp pose. SE(3) equivariance requires that $f(\mathbf{P})=G$ satisfies:
\begin{equation}
f(R \mathbf{P} + t) = R G + t
\end{equation}
for any arbitrary rotation $R\in SO(3)$ and translation $t\in \mathbb{R}^3$. This is often achieved by centering the point cloud $\mathbf{P}$ around its mean, $\frac{1}{n}\sum_i \mathbf{P}_i$, to obtain translation-invariant input. Then, any SO(3) equivariant neural network, such as Vector Neurons (VNN) \cite{Deng2021VectorNA}, is utilized. Finally, the output is re-centered by adding the mean of the point cloud to the output. In this work, we adopt a similar approach to sample grasp poses, but we use the grasp pose representation introduced by M. Sundermeye et al. \cite{Sundermeyer2021ContactGraspNetE6}, which is based on contact points. By utilizing this, we can not only train efficiently but also infer efficiently, giving us an advantage in dynamic grasping.

\subsection{Gaussian Distribution on Riemannian manifold}
\label{sec:riemmanian}
To generate continuous distributions of SE(3) poses, which we require in both the grasp pose post-filtering stage (Section \ref{sec:reachablespace_gmm}) and the task-space reactive control formulation (Section \ref{sec:task_control}), we use the Riemannian Expectation-Maximization (EM) algorithm by Calinon \cite{Calinon2019GaussiansOR}, suitable for Riemannian manifolds like SE(3) with a left-invariant metric. The Gaussian distribution on Riemannian manifold $\mathcal{M}$ is:
\begin{equation}
    \mathcal{N}_\mathcal{M}(x|\mu,\Sigma)=((2\pi)^d |\Sigma|)^{-\frac{1}{2}}e^{-\frac{1}{2}\text{Log}_\mu(x) \Sigma^{-1} \text{Log}_\mu(x)}
\end{equation}
Here, $\mu\in\mathcal{M}$ is the origin of the tangent space, $x\in\mathcal{M}$ is a point on the manifold, $\Sigma\in \mathcal{T}\mu \mathcal{M}$ is a covariance in the tangent space of $\mu$, and $d$ is the dimension of the tangent space (6 for SE(3)). The mean $\mu$ of $N$ points on the manifold is iteratively computed as:
\begin{equation}
    u=\frac{1}{N}\sum_{i=1}^N \text{Log}_\mu (x_i), \quad \mu=\text{Exp}_\mu(u)
\end{equation}
After convergence, the covariance $\Sigma$ is computed as:
\begin{equation}
    \Sigma=\frac{1}{N}\sum_{i=1}^N \text{Log}_\mu (x_i) \text{Log}_\mu^T (x_i)
\end{equation}
Note that the logarithmic and exponential operations in the Riemannian manifold are different from those in the Lie group for SE(3):
\begin{equation}
    \text{Log}_\mu(x) = [\text{Log}_\mathcal{G}(\mu^{-1}x)]^{\vee}_\mathcal{G} = v \in \mathfrak{se}(3)
\end{equation}
\begin{equation}
    \text{Exp}_\mu(v) = \mu \, \text{Exp}_\mathcal{G}([v]^{\wedge}_\mathcal{G}) = x \in \text{SE(3)}
\end{equation}
$\text{Log}_\mathcal{G}$ and $\text{Exp}_\mathcal{G}$ are a logarithmic and exponential operation of the Lie group, converting SE(3) matrix into screw axis matrix representation and vice versa. $[\,]^{\vee}_\mathcal{G}$ and $[\,]^{\wedge}_\mathcal{G}$ are operations that change the screw axis matrix to screw axis compact vector form and vice versa in the Lie algebra.

\section{Method}
Following we detail each step in our feasible dynamic grasping pipeline outline in Figure \ref{pipeline}.
\subsection{Shape Completion with Distance Field Refinement}
\label{sec:distance_field}

\begin{figure}[!tbp]
  \tabcolsep=0.\linewidth
  \divide\tabcolsep by 4
  \begin{tabular}{cc}
    \includegraphics[width=0.5\linewidth]{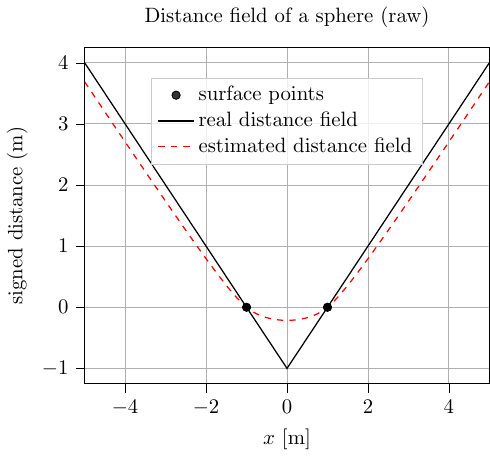} &
    \includegraphics[width=0.5\linewidth]{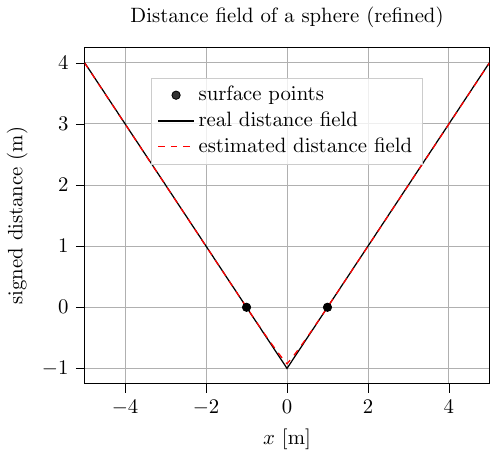} \\
  \end{tabular}
  \caption{A 1D slice of the distance field of a sphere point cloud. The image on the left shows a comparison between the real SDF and the initial estimation from the GPDF. On the right, the image shows an improved estimation after five iterations of gradient descent, employing the ray marching concept. \label{gpds_iteration}}
\vspace{-10pt}
\end{figure}

\begin{figure}[tbp]
  \centering
  \includegraphics[trim={0cm 0.95cm 0cm 1cm},clip,width=\linewidth]{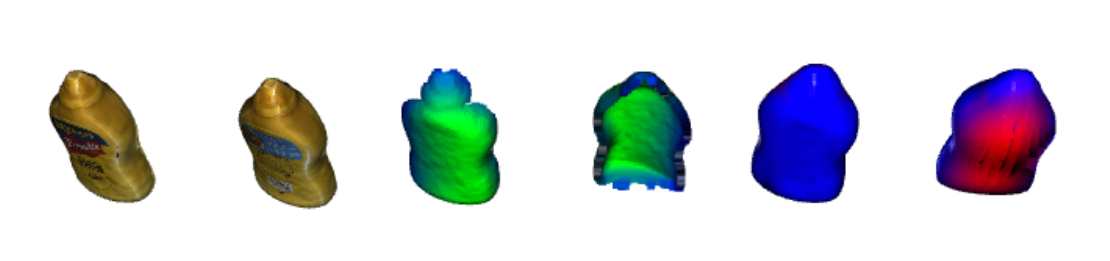}
  \caption{The reconstruction is done using the Gaussian process. The initial two images on the left display the mesh of a mustard bottle. In the subsequent two images, partial point clouds of the object are displayed. Finally, the last image shows the reconstructed point cloud of the object. Areas with low shape uncertainty are colored blue, while regions with high shape uncertainty are highlighted in red.\label{pcd_reconstruction}}
  \vspace{-15pt}
\end{figure}

The Matern $\nu=1/2$ kernel, given by $k_o(d)=\sigma^2 \exp\left(\frac{-d}{l}\right)$ with $\sigma=1$, provides an approximate signed distance function (SDF) for an object, where $l$ controls interpolation degree. This result is due to GP's interpolation abilities and the unique property of the Matern $\nu=1/2$ kernel, which provides an analytic solution for $d$. Other kernels lack analytic inverses or yield non-bijective outputs like $\sqrt{d^2}$. Despite its capability to estimate the sign of the distance, its accuracy diminishes near and inside the object, as in  Figure \ref{gpds_iteration} (left). To enhance accuracy, we employ an iterative refinement inspired by sphere marching.This process relies on two assumptions: 1) The estimated distance field can be represented as a composite function, $d(\mathbf{x})=g(\text{SDF}(\mathbf{x}))$, where $g:\mathbb{R}\rightarrow\mathbb{R}$ is a monotonically increasing function that goes through the origin and $\text{SDF}$ denotes the actual signed distance field; 2) The absolute value of the estimated distance is less than or equal to the actual distance, i.e., $|d(\mathbf{x})|\leq|\text{SDF}(\mathbf{x})|$. The iterative process involves projecting a point $\mathbf{x}$ onto the object's surface using:
\begin{equation}
    \mathbf{x}\leftarrow \mathbf{x} - d(\mathbf{x}) \frac{\nabla d(\mathbf{x})}{||\nabla d(\mathbf{x})||_2}
\end{equation}
Accumulating $d(\mathbf{x})$ values throughout the iterations yields a refined estimate of the distance from the original $\mathbf{x}$. The normalized gradient of the distance field becomes the updated point's surface normal. This iterative process converges without overshooting due to the constant Eikonal function property of the signed distance function, ensuring:
\begin{equation}
    \begin{aligned}
        \frac{\nabla d(\mathbf{x})}{||\nabla d(\mathbf{x})||_2} = \frac{g' \nabla \text{SDF}(\mathbf{x})}{||g' \nabla \text{SDF}(\mathbf{x})||_2}\\
        = \frac{g' \nabla \text{SDF}(\mathbf{x})}{g'||\nabla \text{SDF}(\mathbf{x})||_2} = \nabla \text{SDF}(\mathbf{x})
    \end{aligned}
\end{equation}
If the second condition is met, the process prevents overshooting and converges, as seen in Figure \ref{gpds_iteration}. We determine the best parameter $l$ for GP using the Acronym dataset \cite{Eppner2020ACRONYMAL}, generating partial point clouds for 8872 objects. For each $l$ value, we create a GPDF and calculate the average distance between the complete object's point cloud and the GP-derived surface, akin to Chamfer distance. A small $l$ results in a large mean distance, indicating poor reconstruction of unseen object parts. Conversely, an excessively large $l$ causes over-interpolation, losing high-curvature details. Although adaptive $l$ selection based on point cloud size is possible, we fix it at $l=0.3 m$ through a grid search ($l=[0,0.6]$). This process aids in obtaining a rough object shape, as shown in Figure \ref{pcd_reconstruction}. Unlike DeepSDF \cite{Park2019DeepSDFLC}, GPDF lacks semantic object information but offers versatility for both objects and scenes.

\vspace{-2.5pt}
\subsection{Data-driven grasp sampler}
\label{sec:grasp_sampler}
\begin{figure}
  \centering
  \includegraphics[width=\linewidth]{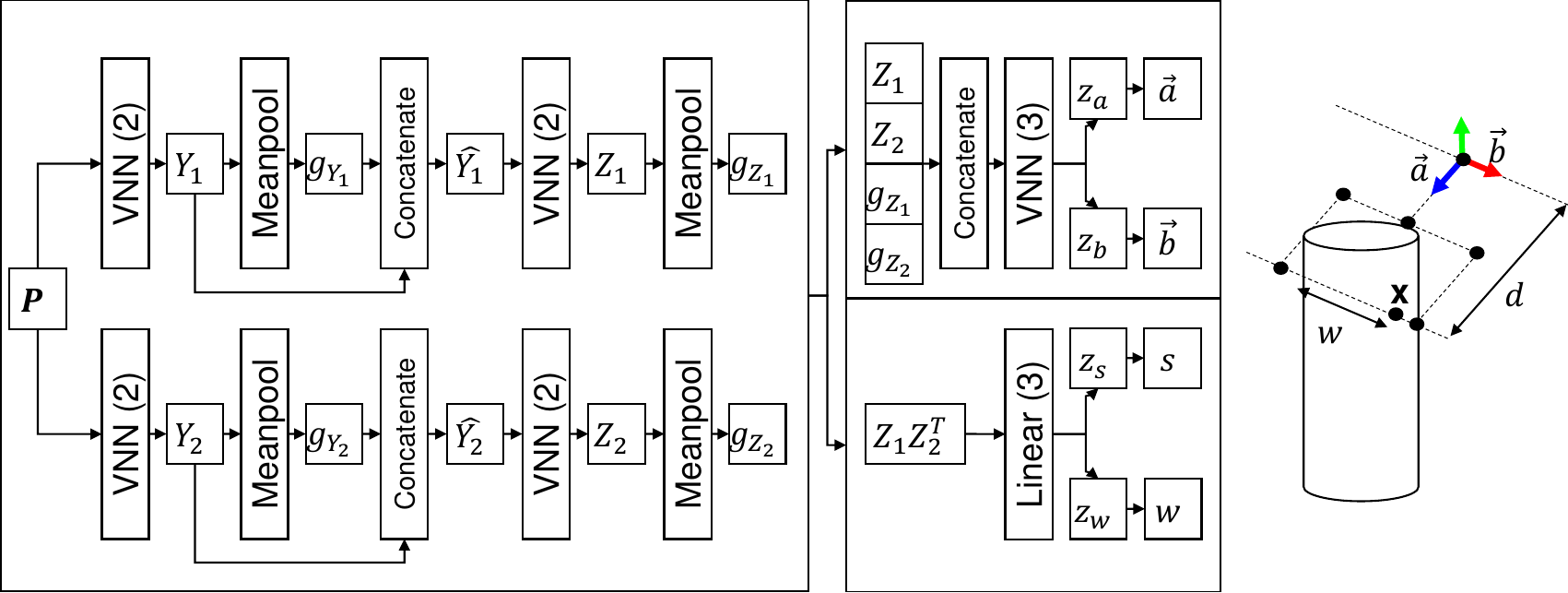} \\
  \caption{Architecture of the data-driven grasp sampler, numbers (*) denote the respective number of layers in each component.\label{nn_architecture}}
  \vspace{-15pt}
\end{figure}
We use two VNN PointNet encoders, one VNN decoder for SO(3) equivariant features and one linear decoder for SO(3) invariant features, as depicted in Figure \ref{nn_architecture}. The input to our neural architectiure, denoted as $\mathbf{P}\in\mathbb{R}^{n\times2\times3}$, includes surface points $\mathbf{x}_{s}\in\mathbb{R}^{n\times3}$ and their surface normals $\nabla d(\mathbf{x}_{s})\in\mathbb{R}^{n\times3}$ extracted from GP. Surface points are generated through the creation of an 8x8x8 grid within the expanded bounding box of the partial point cloud, centered at its midpoint. This configuration can be adjusted according to the camera's orientation. The network's primary objective is twofold: first, to determine whether a given point in $\mathbf{x}_s$ is a potential contact point on one side of the gripper; second, if viable, to estimate two columns of the rotation matrix and the width between the contact points. To validate if this grasp pose estimation is feasible, we get points on the gripper, transform points based on the grasp pose, and then input them into GP to verify whether they lie outside the object's surface. Once we ascertain that all points are indeed situated outside the object, we proceed to retrieve the surface normals of the paired contact points $\mathbf{x}$ and $\mathbf{x}'=\mathbf{x}-w\Vec{b}$ from GP. Subsequently, we examine whether the cosine similarity between the surface normals of these contact pairs closely approximates -1. This similarity metric serves as an indicative measure of force closure for jaw grippers.
\vspace{-2.5pt}
\subsection{Reachable-Space Riemannian Mixture Model (RMM)}
\label{sec:reachablespace_gmm}
Reachability serves as a crucial metric for a robot manipulator to assess the feasibility of reaching a desired workspace pose $G\in SE(3)$ with a given joint configuration $q\in\mathbb{R}^n$. It has various applications, such as guiding mobile manipulators to adjust their legs or wheels \cite{Jauhri2022RobotLO, Vahrenkamp2013RobotPB} and optimizing grasp poses for fixed manipulators \cite{Kim2014CatchingOI, Akinola2021DynamicGW}. Kim et al. \cite{Kim2014CatchingOI} proposed a method utilizing a Gaussian Mixture Model (GMM) to pre-learn reachability, which expedited the process by avoiding explicit inverse kinematics computations with little loss in accuracy. This was accomplished by utilizing end-effector samples from a dataset of joint configurations, thereby enabling the optimization of end-effector poses in conjunction with grasp pose GMM. They fitted GMM by flattening out the first two columns of the rotation matrix and translation, which gave better results than using Euler angles or quaternions. In this work, we instead learn the reachable space based on the Riemannian Gaussian distribution introduced in Section \ref{sec:riemmanian} for better modeling. To build an offline reachable-space Riemannian Gaussian Mixture Model (RMM), we use forward kinematics to derive end-effector poses. From 10 million joint configurations, we downsample to about 3 million, excluding self-collisions and poses below 0.1 m above the table. This will prevent the selection of grasp poses that cause collisions with the table. We determine 64 clusters using the Bayesian Information Criterion (BIC) with K-means++ initialization. Poses with probabilities below the 99\% threshold are considered unreachable. If the environment is more cluttered one could fit a GPDF to the surface boundary and use it to filter out poses. 


\begin{figure}
  \centering
  \includegraphics[width=1.0\linewidth]{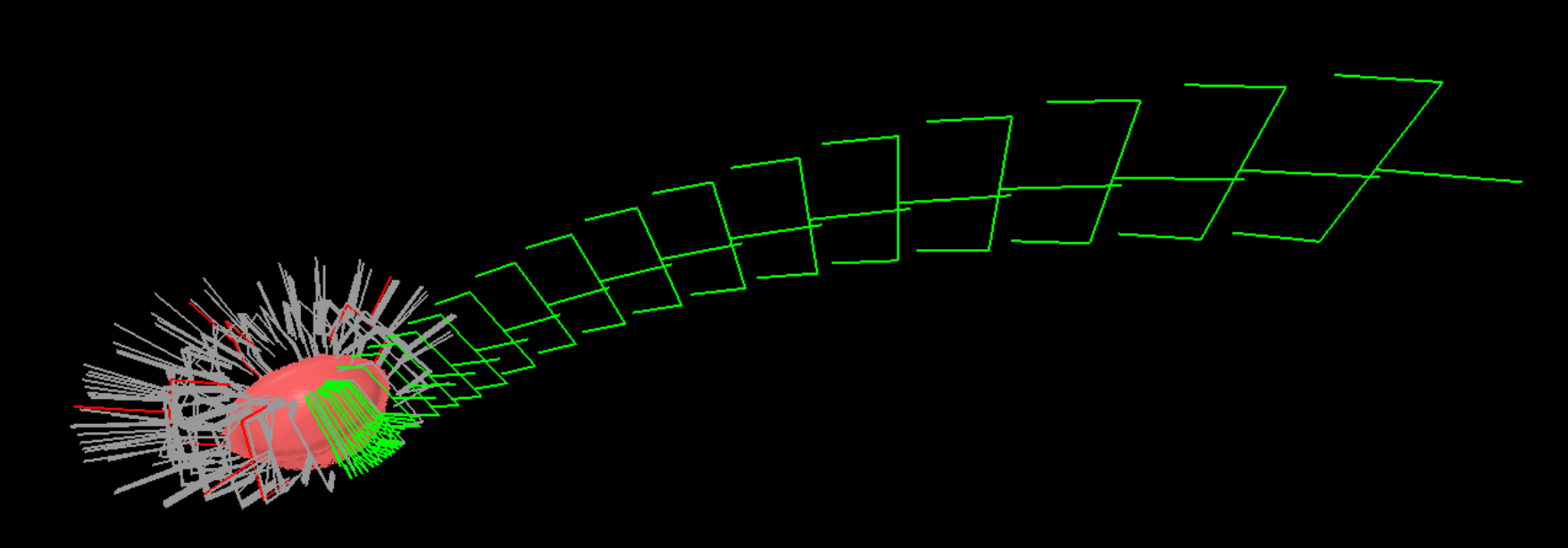}
  \caption{Gray grasp poses represent the outputs from the grasp sampler, the four red grasp poses correspond to clusters' means from the RMM. The green trajectory of the end-effector illustrates a gradient ascent outcome, converging towards a feasible point.\label{gmm}}
  \vspace{-15pt}
\end{figure}
\subsection{Continuous Grasp Pose Distributions for Control}
After filtering grasp poses, multiple viable ones may remain. Due to discrete samples, establishing stable control gradients amid disturbances is challenging. We propose two control strategies based on continuous distribution of grasp poses: task-space grasp representation with RMM and joint-space grasp representation with GPDF. 

\subsubsection{Task-Space Grasp Distribution}
\label{sec:task_control}
RMM is applied to sampled grasp poses, prioritizing more plausible ones. This allows computation of the gradient for the robot's end-effector pose via gradient ascent on the RMM and differential inverse kinematics. In the case of RMM for SE(3), the probability of pose $x$ is given by:
\begin{equation}
    \begin{aligned}
        \mathcal{P}(x|\mu,\Sigma) = \sum_{i=1}^n \pi_{i}
        \mathcal{N}_\mathcal{M}(x|\mu_i,\Sigma_i)
    \end{aligned}
\end{equation}
where $\pi_i$, $\mu_i$, and $\Sigma_i$ are weight, mean, and covariance of cluster $i$. The gradient is computed as follows:
\begin{equation}
    \begin{aligned}
        \frac{\partial\mathcal{P}(x|\theta)}{\partial x}= & \sum_{i=1}^n \pi_{i} \mathcal{N}_\mathcal{M}(x|\theta_i) J_r^{-1}(\mu^{-1})(-\Sigma^{-1}\text{Log}_{\mu_i}(x))
    \end{aligned}
\end{equation}
with $\theta_*=\{\mu_*,\Sigma_*\}$ and $J_r$ is the right Jacobian of SE(3) \cite{Chirikjian2012StochasticMI}. Since obtaining $J_r$ analytically can be challenging, we rely on numerical methods for gradient computation. Subsequently, we use this gradient to compute joint-space velocity controller:
\begin{equation}
\label{eq13}
\Dot{q} =J(q)^{\dagger}\frac{\partial\mathcal{P}(x|\theta)}{\partial x}
\end{equation}
where $J(q)$ is the manipulator Jacobian and $^{\dagger}$ denotes the Pseudo-Inverse. Eq. \ref{eq13} acts as gradient ascent to the closest grasp pose (local maxima of GMM) from the robot's current pose (Figure \ref{gmm}). However, there are two problems with using raw GMM gradients: zero gradient in local minima and zero gradient for poses distant from grasp poses. To escape from the local minima, we sample surrounding points and add a disturbance if they are identified as local minima. To solve the second problem, we use log probability instead, 
\begin{equation}
    \resizebox{\linewidth}{!}{$\begin{aligned}
        &\frac{\partial\log(\mathcal{P}(x|\theta))}{\partial x}=\frac{\sum_{i=1}^n \pi_{i} \mathcal{N}_\mathcal{M}(x|\theta_i) J_r^{-1}(\mu^{-1})(-\Sigma^{-1}\text{Log}_{\mu_i}(x))}{\sum_{i=1}^n \pi_{i} \mathcal{N}_\mathcal{M}(x|\theta_i)}
    \end{aligned}$}
\end{equation}
This gradient offers a numerical advantage when the probability is low, as multiplying a constant to all $\mathcal{N}_\mathcal{M}(x|\theta_i)$ yields the same results. For dynamic objects, we adjust means with the object and compute the gradient. With new samples, we iteratively update the RMM. To ensure reachability, we transform means and refit with reachable grasp poses.

\begin{figure}[!tbp]
  \tabcolsep=0.\linewidth
  \divide\tabcolsep by 6
  \begin{tabular}{ccc}
    \includegraphics[width=0.33\linewidth]{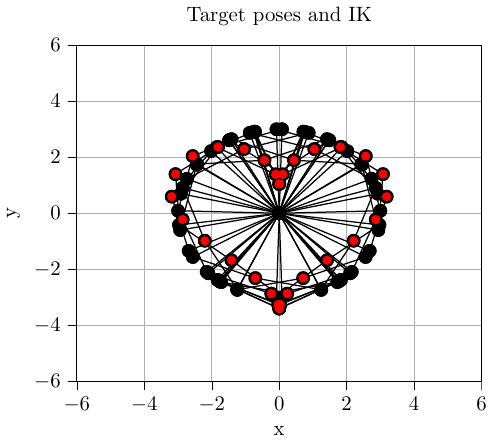} &
    \includegraphics[width=0.33\linewidth]{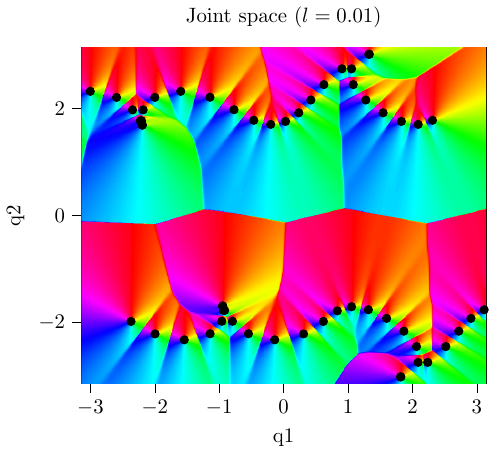} &
    \includegraphics[width=0.33\linewidth]{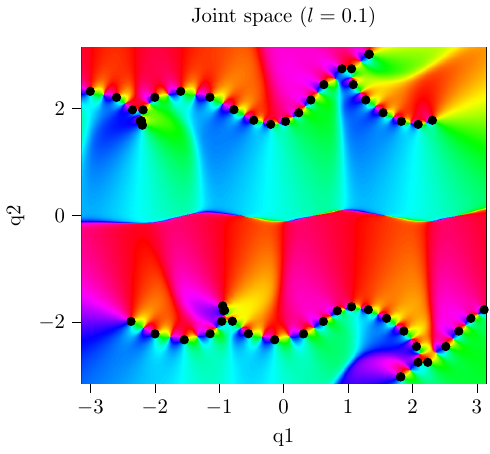}\\
  \end{tabular}
  \vspace{-5pt}
  \caption{Joint-space gradient field derived from GPDF using a 2D robot's IK solutions. (left) End-effector targets as red points, accompanied by their respective IK solutions. (center-right) joint-space gradient field with varying interpolation value $l$.\label{gpds_joint}}
  \vspace{-15pt}
\end{figure}

\subsubsection{Joint-Space Grasp Distribution}
\label{sec:joint_control}
Alternatively, if we have an inverse kinematics (IK) solver for a $n$-DoF robot with joint limits that can yield diverse solutions of $q\in\mathbb{R}^n$ for a set of end-effector poses, we can create a $n$-D distance field using GPDF on the IK solutions. As GPDF is a monotonically increasing function it can be considered  a potential function with multiple equilibria. We thus propose the following joint-space velocity controller,
\begin{equation}
\Dot{q}=-d_q(q)\nabla d_q(q),
\end{equation}
obtained from the GPDF fitted on the IK solutions, $d_q$, as shown in Figure \ref{gpds_joint} for 2-DoF example. Since sampling multiple IK solutions within a short time frame is challenging, we use a combination of the learning-based method, IKFlow \cite{Ames2021IKFlowGD}, and the batched-damped pseudo-inverse method with batched forward kinematics and Jacobian calculation \cite{Zhong_PyTorch_Kinematics_2023}. IKFlow gives multiple IK solutions that already, considering self-collision and joint limits at nearly 100 Hz using normalizing flow. Although its solutions may lack accuracy compared to the pseudo-inverse method, we initially sample IK solutions with IKFlow and refine them with the pseudo-inverse method for 3 iterations. For dynamic objects, we adjust grasp poses based on the object pose or get new grasp poses and resample IK solutions with the same seed for IKFlow, ensuring smooth transitions from previous solutions. To ensure reachability, we discard unreachable grasp poses.

\vspace{-2.5pt}
\section{Experiments}
\vspace{-2.5pt}
\subsection{Experimental Setup}
Our setup includes a KUKA iiwa 7 manipulator with a SAKE gripper and a Realsense D435 depth camera, utilized in both real robot and simulation experiments. Both task and joint space grasp representations are executed on an Intel i7-11700K without a GPU. However, for comparison, the grasp diffusion model runs on an Nvidia RTX 3080 GPU. Our primary objectives are twofold: firstly, to evaluate the quality of grasp poses generated by the grasp sampler pipeline, and secondly, to assess the feasibility of dynamic object grasping using joint and task space representations.

To evaluate grasp poses, we conduct experiments in both simulation and on a real robot, recognizing the potential dynamic nature of objects. Initially, experiments assume static objects with one-shot grasp sample predictions due to practical constraints, such as on-hand camera limitations where objects might go out of sight. We acknowledge the necessity of addressing dynamic scenarios in future research.

\subsubsection{Grasp Sampler Test}
To evaluate grasp poses, we conduct experiments in both simulation and on a real robot, assuming that the object remains static. In both simulation and real-world experiments, we utilize the YCB datasets \cite{alli2015BenchmarkingIM}, ensuring that the data used for evaluation were not part of the training dataset for the data-driven grasp sampler.

In simulation, we use Pybullet with fine-tuned friction coefficients, testing with 50 different YCB objects. Objects are initialized at random poses within a 10cm cube located at (x: 40cm, y: 0cm, z: 60cm) relative to the robot's base coordinate. Our methodology involves capturing segmented RGBD images, sampling and filtering grasp poses, employing task-space representation for stability assessment, executing grasps, applying gravity simulation, and transitioning to a lifting pose. For benchmarking, we compare our approach against the SE(3) grasp diffusion model \cite{Urain2022SE3DiffusionFieldsLS}, which takes a partial point cloud as input unlike neural grasp distance \cite{Weng2022NeuralGD}, and an alternative method of sampling 100 discrete grasp poses from the grasp diffusion model and using our task-space representation. We justify the comparison with the SE(3) grasp diffusion model due to its prediction of the signed distance function (SDF) of the object and utilization of continuous grasp representation.

The success criterion is determined by whether the object remains stable for at least 2 seconds after the robot transitions to the lifting pose. This criterion aims to capture not only the initial grasp success but also the ability to maintain stability during subsequent manipulation steps. And we do this 10 times for each object. For the real robot experiment (Figure \ref{realrobot}), we provide a segmented RGBD image to the robot using the method described in \cite{mobile_sam}, sample grasp poses, convert them to joint space grasp representations, and assess the robot's ability to successfully grasp the objects.
\begin{table}[!tbp]
\centering
\caption{Grasp Success Rate for Static Objects in Simulation}
\label{tab:grasp_quality}
\resizebox{\columnwidth}{!}{%
\begin{tabular}{|c|c|}
\hline
\textbf{Method}                                                & \textbf{Accuracy (\%)} \\ \hhline{|=|=|}
Grasp diffusion \cite{Urain2022SE3DiffusionFieldsLS} + Energy function                     & 67.23 (29.59) \\ \hline
Grasp diffusion \cite{Urain2022SE3DiffusionFieldsLS} + Task-Space distribution             & 72.46 (19.84) \\ \hhline{|=|=|}
Grasp sampler (ours) + Task-Space distribution        & 77.60 (24.54) \\ \hline
(*) + Post-filtering                                      & 88.60 (15.75) \\ \hline
(*) + Reachability analysis                               & \cellcolor{blue!15}  \textbf{90.00 (11.49)} \\ \hline
\end{tabular}%
}
\vspace{-15pt}
\end{table}

\begin{table}[!tbp]
\caption{Comparison for Dynamic Object Grasping in Simulation}
\vspace{-2.5pt}
\resizebox{\columnwidth}{!}{%
\begin{tabular}{|c|c|c|c|c|c|}
\hline
\textbf{Trajectory} & \begin{tabular}[c]{@{}c@{}}\textbf{Method} \end{tabular} & \begin{tabular}[c]{@{}c@{}}\textbf{Self} \\ \textbf{collision (\%)} $\downarrow$ \end{tabular} & \begin{tabular}[c]{@{}c@{}}\textbf{Joint limit}\\ \textbf{violations (\%)} $\downarrow$ \end{tabular} & \begin{tabular}[c]{@{}c@{}}\textbf{Manipul} \\ \textbf{-ability}\end{tabular} $\uparrow$ & \begin{tabular}[c]{@{}c@{}}\textbf{Gripper-object} \\ \textbf{distance (cm)}$\downarrow$\end{tabular} \\ \hline
\multirow{5}{*}{\textbf{Circular}} & \begin{tabular}[c]{@{}c@{}} Joint \\ space (ours) \end{tabular} & \cellcolor{blue!15} \textbf{1.08 (0.57)} & \cellcolor{blue!15} \textbf{57.06 (6.52)} & \cellcolor{blue!15} \begin{tabular}[c]{@{}c@{}} \textbf{0.0936} \\ \textbf{(0.0017)} \end{tabular} & 16.65 (0.96) \\ \cline{2-6} 
 & \begin{tabular}[c]{@{}c@{}} Task \\ space (ours) \end{tabular} & 6.72 (5.55) & 89.64 (0.74) & \begin{tabular}[c]{@{}c@{}} 0.0729 \\ (0.0022) \end{tabular} & \cellcolor{blue!15} \textbf{3.31 (0.25)} \\ \cline{2-6}
 & \begin{tabular}[c]{@{}c@{}} Grasp \\ diffusion \cite{Urain2022SE3DiffusionFieldsLS} \end{tabular} & 38.14 (11.37) & 95.47 (0.04) & \begin{tabular}[c]{@{}c@{}} 0.0766 \\ (0.0025) \end{tabular}& 17.63 (0.79) \\  \hline
\multirow{5}{*}{\textbf{Linear}} & \begin{tabular}[c]{@{}c@{}} Joint \\ space (ours) \end{tabular} & \cellcolor{blue!15} \textbf{0.00 (0.00)} & \cellcolor{blue!15}  \textbf{55.49 (5.34)} & \begin{tabular}[c]{@{}c@{}} 0.0887 \\ (0.0052) \end{tabular}& 6.44 (0.24) \\ \cline{2-6} 
 & \begin{tabular}[c]{@{}c@{}} Task \\ space (ours) \end{tabular} & 3.84 (2.12) & 62.90 (30.58) & \begin{tabular}[c]{@{}c@{}} 0.0785 \\  (0.0076) \end{tabular}& \cellcolor{blue!15}  \textbf{2.70 (0.19)} \\ \cline{2-6} 
 & \begin{tabular}[c]{@{}c@{}} Grasp \\ diffusion \cite{Urain2022SE3DiffusionFieldsLS} \end{tabular} & 5.60 (0.77) & 86.91 (0.89) & \cellcolor{blue!15}\begin{tabular}[c]{@{}c@{}} \textbf{0.0942} \\  \textbf{(0.0035)} \end{tabular} & 16.12 (0.55) \\ \hline
\multirow{5}{*}{\textbf{Sinusoidal}} & \begin{tabular}[c]{@{}c@{}} Joint \\ space (ours) \end{tabular} & \cellcolor{blue!15} \textbf{0.00 (0.00)} & \cellcolor{blue!15} \textbf{55.64 (3.82)} & \cellcolor{blue!15}\begin{tabular}[c]{@{}c@{}} \textbf{0.0884} \\ \textbf{(0.0021)} \end{tabular}& 12.47 (0.45) \\ \cline{2-6} 
 & \begin{tabular}[c]{@{}c@{}} Task \\ space (ours) \end{tabular} & 1.03 (1.97) & 91.63 (5.63) & \begin{tabular}[c]{@{}c@{}} 0.0847 \\ (0.0033) \end{tabular} & \cellcolor{blue!15} \textbf{7.10 (0.77)} \\ \cline{2-6} 
 & \begin{tabular}[c]{@{}c@{}} Grasp \\ diffusion \cite{Urain2022SE3DiffusionFieldsLS} \end{tabular} & 28.88 (3.99) & 89.36 (1.17) & \begin{tabular}[c]{@{}c@{}} 0.0754 \\ (0.0032) \end{tabular} & 14.85 (1.36) \\ \hline
\end{tabular}%
}
\label{tab:comparison}
\vspace{-15pt}
\end{table}

\subsubsection{Dynamic Object Test}
To assess the robot's ability to grasp a moving object in simulation, we create circular, linear, and sinusoidal trajectories with constant object rotation (see accompanying video). We repeat testing five times for each trajectory, evaluating joint limit violation, manipulability ($m=\sqrt{\det(J(q)J(q)^T)}$), and self-collision without constraints. In real-world tests (Figure \ref{realrobot}), we segment the RGBD image of the object, track its movement relative to the robot's base frame using an Optitrack tracking system, and control the robot based on sampled grasp poses to verify grasp success. For grasp diffusion model, we use energy function directly to calculate the end-effector gradient.
\vspace{-5pt}
\subsection{Experimental Results and Discussion}
\vspace{-2.5pt}
\subsubsection{Grasp quality}
In simulation, we achieved a 90.00\% success rate in grasping out of 500 trials involving 50 distinct YCB objects, as detailed in Table \ref{tab:grasp_quality}. Without post-filtering, grasp success significantly dropped, emphasizing the need for validating sampled grasp poses. Initially, the grasp diffusion model had poor results using an energy function due to non-geodesic trajectories in task space with gradient descent. Substituting the energy function with our task-space distribution improved performance but slowed inference to over 10 seconds, much longer than the 58 Hz with the energy function. However, even with this adjustment, performance remained inferior to our grasp sampler without filtering.

Our data-driven grasp sampler, running within 10 ms for 1024 points on CPU, may yield inaccurate samples since it is a small model. Full shape information from the distance field helps mitigate these errors. However, this can lead to overly conservative results, offering few grasp poses whereas the diffusion model can create infinite number of samples. Similar to our approach with IKFlow, we can sample IK solutions from data-driven methods and refine them numerically for grasp pose optimization \cite{Turpin2023FastGraspDDM} for improvement. Reachability analysis marginally improves sampler accuracy as the objects were already near the robot, where most grasp poses are reachable. Post-filtering and reachability analysis take less than 5 ms. However, grasp pose optimization becomes a bottleneck due to its $O(n^3)$ complexity, taking 0.25 seconds for 512 points and 1-2 seconds for 1024 points.

In real robot testing with static objects, 14 out of 16 grasps were successful for 8 different objects. Failures occurred with the drill due to its heavy unbalanced weights and the banana due to excessive grasp strength causing slip. Similar failures happened with 19 out of 24 grasps for eight dynamic objects, despite good initial grasp contacts, due to object shape considerations.

\begin{table}[tbp]
\centering
\caption{Control Frequency of Task vs Joint Space Controllers}
\resizebox{\linewidth}{!}{%
\begin{tabular}{|c|c|c|c|c|c|c|c|c|c|c|}
\hline
\textbf{Number of GMM means} & 1 & 2 & 3 & 4 & 5 & 6 & 7 & 8 & 9 & 10 \\ \hline
\textbf{Frequency (Hz)} & 106.6 & 99.4 & 95.7 & 90.2 & 88.6 & 85.2 & 83.1 & 81.1 & 79.4 & 78.2 \\ \hline \hline
\textbf{Number of GP’s samples} & 10 & 20 & 30 & 40 & 50 & 60 & 70 & 80 & 90 & 100 \\ \hline
\textbf{Frequency (Hz)} & 22.4 & 18.3 & 16.7 & 17.2 & 16 & 15.1 & 15.8 & 14.8 & 14.1 & 13.9 \\ \hline
\end{tabular}%
}
\label{tab:frequency}
\vspace{-15pt}
\end{table}

\subsubsection{Comparison between task and joint space methods} 
The task space method surpasses the joint space method in computational efficiency, as demonstrated in Table \ref{tab:frequency}. Specifically, the task space method shows notably faster execution times, rendering it more appealing. Notably, its frequency demonstrates nearly linear decrease with increasing GMM means, typically requiring only 4 to 8 means for adequate object grasp pose representation. In contrast, the frequency of the joint space method doesn't follow a straightforward linear relationship with the number of IK solutions, largely due to parallel computing advantages.

Interestingly, the joint space method can exhibit speed enhancements even with increasing samples, owing to parallel computations and memory allocations at every power of 2. However, achieving smooth control with the joint space method necessitates a substantial number of samples, typically around 100, to ensure gradual joint pose changes and avoid abrupt transitions. From a robot kinematics standpoint, the joint space method outperforms the task space method, as highlighted in Table \ref{tab:comparison}. This superiority arises from the task space method's susceptibility to singularities, while the joint space method can navigate around singularities, joint limits, and self-collision, resulting in higher manipulability scores. The grasp diffusion model often underperforms because it calculates the end-effector gradient similar to the task space method, resulting in an unnatural robot trajectory due to the gradient field it generates. Consequently, the strength of diffusion models primarily lies in offline optimization.

Although it might appear that the joint space method could outperform the task space method solely by achieving higher speeds, this assumption oversimplifies the comparison. For instance, in an HRI scenario like object handovers, the joint space method may be less intuitive for humans, resulting in perceived inefficiency due to larger distances between the gripper frame and the object frame during motions. In our real robot object handover tests, the task space method was preferred in this regard. The suitability of each method depends on the task's requirements and constraints.

\section{CONCLUSION}
In conclusion, our research focuses on enhancing robots' object-grasping abilities in dynamic scenarios. We stress the importance of considering factors like object shape, potential collisions, and robotic motion when devising effective grasping strategies. We introduced a step-by-step approach that incorporates Gaussian Process Distance Field for object shape reconstruction, SE(3) equivariant grasp sampling and filtering, and Riemannian Mixture Models for grasp pose selection and control, aimed at improving robotic grasping efficiency. The proposed method provides a viable solution for generating and evaluating grasp poses, while also facilitating motion planning and control in dynamic environments, with potential applications in more complex settings.

\bibliographystyle{ieeetr}
\bibliography{bibliography}

\end{document}